\def\year{2022}\relax
\documentclass[letterpaper]{article} 
\usepackage[frozencache,cachedir=minted-cache]{minted}
\usepackage{aaai22}  
\usepackage{times}  
\usepackage{helvet}  
\usepackage{courier}  
\usepackage[hyphens]{url}  
\usepackage{graphicx} 
\usepackage{natbib}  
\usepackage{caption} 
\DeclareCaptionStyle{ruled}{labelfont=normalfont,labelsep=colon,strut=off} 
\usepackage{algorithm}
\usepackage{algorithmic}
\usepackage{hyperref}

\usepackage[utf8]{inputenc}

\usepackage{newfloat}
\usepackage{listings}

\usepackage{adjustbox}
\usepackage{multirow}
\usepackage{bbm}
\usepackage{mathtools}
\usepackage{booktabs}
\usepackage{amssymb}
\usepackage{soul}
\usepackage{color}
\usepackage{subfigure}
\usepackage{xcolor}

\floatstyle{ruled}
\newfloat{listing}{tb}{lst}{}
\floatname{listing}{Listing}

\title{Application of Deep Learning in Generating Structured Radiology Reports: \\A Transformer-Based Technique\footnote{Code is available on \url{https://github.com/realsarm/ReportQL}}\footnote{This is a pre-peer-review, pre-copyedit version of this article. The final authenticated version is available online at: \url{https://doi.org/10.1007/s10278-022-00692-x}}}
\author{
    Seyed Ali Reza Moezzi\textsuperscript{\rm 1}, Abdolrahman Ghaedi\textsuperscript{\rm 1}, Mojdeh Rahmanian\textsuperscript{\rm 1}, Seyedeh Zahra Mousavi\textsuperscript{\rm 2}, Ashkan Sami\textsuperscript{\rm 1}\thanks{
    	Corresponding author: see final version
    }
    \\
}
\affiliations{
    \textsuperscript{\rm 1} Department of Computer Science and Engineering and IT, Shiraz University, Shiraz, Iran\\
    
    \textsuperscript{\rm 2} Shiraz University of Medical Sciences, Shiraz, Iran
}

\usepackage{bibentry}

\begin{document}

\maketitle

\begin{abstract}\label{abstract}
Radiology reports needed for clinical practice and research are written and stored in free-text narrations making extraction of relative information for further analysis difficult. In these circumstances, natural language processing (NLP) techniques can facilitate automatic information extraction and transformation of free-text formats to structured data. In recent years, deep learning (DL)-based models have been adapted for NLP experiments with promising results. Despite the significant potential of DL models based on Artificial Neural Network (ANN) and Convolutional Neural Network (CNN), the models face some limitations to implement in clinical practice. Transformers, another new DL architecture, have been increasingly applied to improve the process. Therefore, in this study, we propose a transformer-based fine-grained Named Entity Recognition (NER) architecture for clinical information extraction. We collected 88 abdominopelvic sonography reports written in free-text formats and annotated them based on our developed information schema. The text-to-text transfer transformer model (T5) and Scifive, a pre-trained domain-specific adaptation of the T5 model, were applied in two phases of pre-training and fine-tuning to extract entities and relations and transform the input to a structured format. Our transformer-based model applied in this study outperformed previously applied approaches such as ANN and CNN models based on ROUGE-1, ROUGE-2, ROUGE-L, and BLEU scores of 0.732, 0.633, 0.635, and 0.727, respectively while providing an interpretable structured report. 
\end{abstract}

Keywords: Structured Reporting; Named Entity Recognition; Relation Extraction; Natural Language Processing; Deep Learning; Transformers

\section{Introduction}\label{intro}

A radiology report narrative is a text-based interpretation of the acquired images, containing essential information about the patient's history, imaging procedure, and radiologic findings. However, these reports are mainly written in unstructured, free-text narrations that are hard for computers to interpret and extract clinical information. The variety of medical lexicons used by different radiologists besides the complexity and ambiguity of free-format narrations make the automated analysis more difficult.

Despite these challenges, radiology reports created by different institutions mostly follow the same semantic framework consisting of observations, observation modifiers such as anatomy, quantity, and severity, and negation providing substantial opportunity for natural language processing (NLP) techniques to convert these free-text narrations to structured format \cite{hassanpour2016information}.
Medical NLP experiments mainly consist of two critical steps, named entity recognition and relation extraction. “Named Entity Recognition (NER)" or “entity extraction” is defined as scanning the unstructured texts to find, label, and classify specific terms into categories such as observation, anatomy, location, etc. For instance, in the sentence “a simple cortical cyst is seen in the left kidney”, NER can extract the terms “simple”, “cortical cyst”, and “left kidney” and categorize them in the form of “simple: modifier”, “cortical cyst: clinical finding”, and “left kidney: anatomy” without establishing any relations between them. This step can be followed by “relation extraction” which aims to find meaningful relations among the entities identified through NER tasks. Considering the mentioned example, relation extraction methods can establish a relationship among “simple”, “cortical cyst”, and “left kidney” in the way that “simple” and “left kidney” are the type and anatomical modifiers of the “cortical cyst” \cite{perera2020named, steinkamp2019toward}. However, although in NLP tasks, NER should be accompanied by relation extraction to transform unstructured free texts to a structured format, most studies have focused only on one of these methods and in particular the NER. Structured radiology reporting has been globally advocated as a promising solution to improve the quality and clarity of radiology reports and enable data mining needed for research processes. However, obstacles towards the conversion of free-text to the structured format such as variability in language, length, and writing style have limited its global application.

For years, some traditional NLP techniques such as rule-based and dictionary-based models have been used; however, recent advances in deep learning (DL) algorithms have improved NLP performance \cite{sorin2020deep, monshi2020deep}. Recent studies in the general English domain have reported that the new deep learning architecture called "transformers'' has provided further improvements leading to the global interest for applying transformers in clinical domains \cite{wolf2020transformers, devlin2018bert, lan2019albert, liu2019roberta, raffel2020exploring}. The use of attention models such as transformers may also enable the system to control long-range text dependencies more effectively than Recurrent Neural Network (RNN) models. Several studies have examined the transformer-based models for some clinical domains such as the biomedical data with promising results confirming their practical applicability \cite{lee2020biobert, si2019enhancing, peng2019transfer, alsentzer2019publicly}.

Since the development of structured radiology reports are required and most experiments in this field have focused on clinical text mining approaches rather than providing both physician and machine-readable systematic structured report, in this study, we propose a practical model of fine-grained NER which employs transformers, e.g. T5 \cite{raffel2020exploring} and domain-specific pre-trained versions such as Scifive \cite{raffel2020exploring} as an element of transfer learning to transform radiology free texts to structured reports. Instead of iterating through a document, we use a machine learning model with generating a machine-readable representation of the extracted information utilizing the pre-existing schemas as supervision.

This paper introduces an end-to-end information extraction system that directly utilizes schemas that are generated manually as training data for a text generation model and requires little to no custom configuration and eliminates the need for additional manual annotation or a heuristic alignment step. We jointly learn to extract, select, and standardize values of interest, enabling direct generation of a machine-readable data format.

The rest of this paper is organized as follows. Section 2 provides a brief review of related works. Section 3 describes an information model, the corpus, and the annotation scheme. In Sections 4 and 5, we report and discuss the results of our experiments and their limitations. Finally, Section 6 summarizes and concludes our work.

\section{Related Works}
 In this section, we have a brief review of literature on structured reporting, strategies used for NER and relation extraction as the main steps of NLP, and the application of DL for this purpose.

\subsection{Structured reporting}
Structured reporting was primarily introduced by DICOM in 1999 with generic models of various reports \cite{european2018esr}. However, the models did not receive global adoption. In 2007-2008, the Radiological Society of North America (RSNA) created a structured reporting initiative \cite{langlotz2006radlex}. For this purpose, data mining approaches based on images and clinical texts known as clinical text mining were used \cite{tayefi2021challenges}. To develop image-based structured reports, Keek et al. and Lambin et al. had used "Radiomics" to convert clinical images to quantitative representation for data mining \cite{lambin2017radiomics, keek2018review}. In another approach, Litjens et al. and Lundervold et al. applied deep learning to interpret images and extract relative features \cite{litjens2017survey, lundervold2019overview}.

For clinical text mining, many institutions have focused on developing reporting templates such as the disease-specific reporting templates collected in the RadReport or other cloud-based solutions with proprietary template formats such as the Smart Reporting application \cite{pinto2018big}. In recent years, the conversion of free-text formats to structured data has been advocated. Taira et al. have introduced a natural language processor to automatically structure medical data. For this purpose, NER and relation extraction are also needed \cite{taira2001automatic}.

\subsection{Named entity recognition and relation extraction} 
Early applications used for bio-NER such as cTAKES and Metamap matched text phrases with dictionaries and rules. Clinical Text Analysis and Knowledge Extraction System (cTAKES) was created by Mayo Clinic integrating rule-based and ML-based approaches \cite{savova2010mayo}. Metamap, a widely available application created by the National Library of Medicine (NLM) used specific dictionaries to find concepts in the biomedical texts and mapped them to the UMLS metathesaurus \cite{aronson2010overview}. However, rule-based and dictionary-based applications need radiologist expertise to define the rules and templates and create dictionaries manually. Also, due to the large-scale medical terminologies that exist, one dictionary cannot cover all available medical terms leading to missing entities \cite{sun2018data, rebholz2011assessment}. Recently, these models have been integrated or replaced by Machine Learning (ML)-based models \cite{ji2020bert, soysal2018clamp}. In any case, these models also require annotated datasets for the training phase.

As mentioned, the next step of the NLP process after the NER is relation extraction. Basically, three approaches have been introduced for this purpose; co-occurrence approaches \cite{rebholz2007ebimed, pennington2014glove, hoffmann2005implementing, garten2009pharmspresso}, pattern-based models \cite{hakenberg2010mining}, and machine learning-aided algorithms \cite{muzaffar2015relation}. However, to handle complex text mining and relation extraction processes, systems based on a combination of two approaches or ML and DL-based algorithms were created.

\subsection{Application of DL-based approaches}

DL-based NLP models including convolutional neural networks (CNN), recurrent neural networks (RNN), long short-term memory (LSTM), and transformer-based models have outperformed traditional NLP techniques in recent years. Chen et al. created a deep learning convolutional neural network (DL-CNN) model for medical text processing \cite{chen2018deep}. An automatic text classification method was proposed by Wang et al. in \cite{wang2019clinical}. The results revealed that the CNN model outperformed the rule-based NLP and other ML models in two of three clinical datasets implying that CNN can reveal hidden patterns that are not captured in the rule-based NLP models. Recurrent neural networks and in particular Long short-term memory (LSTM) have also been used for medical text processing. In a study conducted by Lee et al., LSTM-RNN was applied for automatic classification of orthopedic images based on the presence of a fracture \cite{lee2019automatic}. Carrodeguas et al. created an LSTM-RNN model to extract follow-up recommendations from radiology reports \cite{carrodeguas2019use}.

Bidirectional Encoder Representations from Transformers (BERT) models have been recently applied for text processing. In \cite{smit2020chexbert}, a BERT-based model named “CheXbert” was created to perform automatic labeling and identify the presence of specific radiologic abnormalities in the chest X-rays. Wood et al. also introduced an attention-based model for automatic labeling of head MRI images (ALARM) regarding the presence and absence of five major neurologic abnormalities \cite{wood2020automated}. Additionally, transformers and pre-trained language models are used in some other studies for various purposes such as machine translation \cite{vaswani2017attention}, document generation \cite{liu2018generating}, and Text-to-SQL approaches \cite{lyu2020hybrid}.
Overall, although these studies have paved the way for the global adaptation of DL-based models and in particular the transformers in the medical practice, their application in the development of structured radiology reports is less examined. Our research is inspired by works that have been done in the task of Text-to-SQL transformation \cite{lyu2020hybrid}.

\section{Structured Reporting using Transfer Learning}

In this section, similar to what has been done in \cite{lyu2020hybrid}, we also formulate the structured reporting problem as a multi-task learning problem by adapting a pre-trained Transformer model and translating free-format reports to structured representations. We describe how to develop schemas, prepare data, and train tasks that produce structured output.

\subsection{Developing schemas}

Examining a set of available reports, we developed a schema of all the information documented in the radiology reports. For feasibility studies, we used only abdominopelvic radiology reports to limit possible information domains. However, most information schemas can be applied to all parts of the body.

\begin{figure*}[t]
	\centering
	\includegraphics[width=13cm]{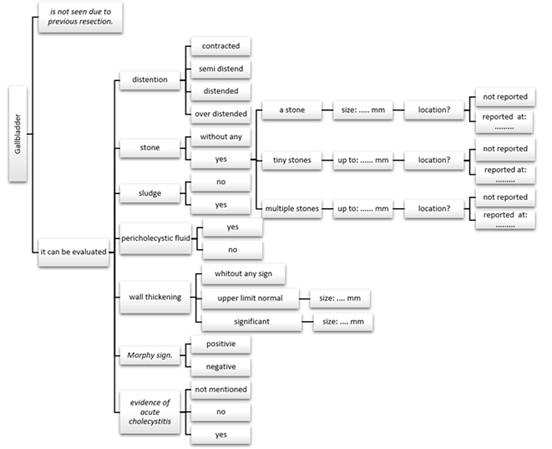}
	\caption{An example of the information schema designed as a hierarchical diagram summarizing the information that can be reported for the gallbladder organ in sonography reports including clinical statements or facts and modifiers such as location or quantity.}
	\label{fig:2}
\end{figure*}

For this purpose, two experts in the field of radiology were asked to extract this information. The main unit of information in our design is a clinical statement, statement, or "fact," such as "a radiological finding was observed". This modifying range includes common linguistic elements such as negation, uncertainty, and conditionality, as well as factual information (e.g., a radiological finding has "location", "size", and “description"). Figure \ref{fig:2} shows an example of a diagram designed for the Gallbladder.

\subsection{Datasets}

For our dataset, we used a practical sample of 88 abdominopelvic radiology reports from people who had made a consensus that their medical information could be used for medical research. The reports were obtained from Namazi and Faghihi Hospitals, the affiliated medical centers of Shiraz University of Medical Sciences, Shiraz, Iran in January 2020. We eliminated all the personal and gender-specific information. The reports included adult patients of all ages and genders written by various attending radiologists and resident radiologists using their personal reporting templates without a unified format and headings and were written in prose.

\begin{figure*}[t]
	\centering
	\includegraphics[width=13cm]{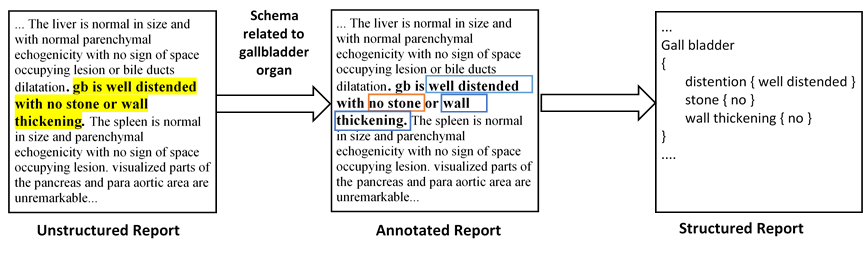}
	\caption{An example of the free text sonography report with the annotated text for the gallbladder organ and the final expected structured report for the gallbladder.}
	\label{fig:3}
\end{figure*}

Data annotation was performed with two annotators in two steps. In the first step, multiple reports were assigned to each annotator. They read reports and based on schemas, the value was assigned to a corresponding key in the schema (see Figure \ref{fig:3}). The second step is cross-validation, in which the annotated data from one annotator was cross-checked by the other annotator. The agreement of the two annotators calculated by Cohen's kappa was 0.86. This shows that the annotators had a high degree of agreement in annotating data.

\subsection{Transformer models}

One of the main challenges of our research was the limitation of training data. To address this problem, we came up with the idea of using transfer learning. A transfer learning model typically consists of two stages: first, unsupervised training on a general-purpose corpus and then, supervised training for a specific task known as the downstream task. In the first stage, the transformer-based model is optimized with a large amount of unannotated text data through language modeling techniques that are independent of specific downstream NLP tasks. In the second stage, the pre-trained transformer-based model is fine-tuned for a particular NLP task using a supervised approach.

We leverage pre-trained language models such as BERT in NLP tasks. BERT is not a unified transfer learning method because BERT-style models can only produce a single prediction for a given input. These models are simply not designed for text generation tasks such as question-answering or summarization. The text-to-text transfer transformer (T5) model overcomes this limitation by outputting a string of text for each input, allowing for both question-answering, summarization, and other tasks where a single output is generally insufficient. In our work, we applied the T5 model and Scifive which is a pre-trained, domain-specific adaptation of the T5 model that is intended for tasks relating to biomedical literature which is very important for the learning process.

\begin{itemize}
	\item \textbf{T5}: The Text-to-Text Transfer Transformer (T5) trained on the "Colossal Clean Crawled Corpus" (C4) uses a basic encoder-decoder transformer architecture as originally proposed by Vaswani et al. \cite{vaswani2017attention}. Each encoder block consists of a self-attention layer and a feed-forward neural network. Each decoder block consists of a self-attention layer, an encoder-decoder attention layer, and a feedforward neural network. T5 is pre-trained on masked language modeling with a learning objective called “span masking”, where spans of text are replaced with a mask token randomly and the model is trained to predict the real masked spans.
	
	The T5 has five different model sizes – Small (60 million parameters), Base (220 million), Large (770 million), 3B, and 11B.
	
	\item \textbf{Scifive}: Like T5, Scifive follows a text-to-text encoder-decoder architecture that transforms an input sequence into an output sequence. Scifive is a domain-specific adaptation of the T5 model pre-trained on a large biomedical corpus. Two different corpora of biomedical language are utilized to generalize the model within the biomedical domain \cite{phan2021scifive}:
	
	\begin{itemize}
		\item PubMed Abstract: It contains over 32 million citations and abstracts of biomedical literature.
		\item PubMed Central: This database is a corpus of free full-text articles in the domain of biomedical and life sciences.
	\end{itemize}
	
\end{itemize}

\subsection{Input representation}

As mentioned, we formulate our task as a text-to-SQL problem. Since the fine-grained structure is a complicated task and hard to reach consensus on specific terms, to remove the burden of this we provided a separate schema to our model which acts as the context for our model as explained in section 3.1. This is fairly similar to a recent text-to-SQL architecture where the schema of the tables is concatenated along with input.  From our formulation, each input includes the schemas and a document.

Scarce of datasets in medical areas, unlike general tasks like question-answering, summarization, etc. may lead to poor performance. To mitigate this issue, we first trained our model on a biomedical question-answering task to further enhance its generalization power. Moreover, we used the masked-language-model technique to train our model using an unsupervised objective. We use one of the strategies explored in the original T5 paper \cite{raffel2020exploring} to establish an unsupervised object to improve the discerning power of our model. To generate training data of sequence span rewriting in a self-supervised fashion, we first randomly replaced tokens of arbitrary length in a natural language medical report with a mask token. The model is fed with this masked token sequence, and trained to produce the original sequence. Masking is governed by the corruption rate parameter which in our findings corruption rate of 15\% works best on generalizing power.

\subsection{Pre-training and fine-tuning of transformer models}

The training process of our model was done in two stages: (i) pre-training and (ii) fine-tuning. For the first stage, the pre-trained model on domain-specific data was reused. Hyperparameters were the same among our experiments. To train and evaluate the proposed model, we randomly separated 80\% of reports as the training set, and 20\% as the test set. Models were trained until validation loss increased. The performance of each of our models was evaluated using ROUGE (ROUGE-1, ROUGE-2, and ROUGE-L) in addition to SacreBLEU and BLEU scores. Another metric utilized was exact-match, considering the number of key-value pairs matching their reciprocal targets. For the second phase of pre-training, our model was trained on 80 hand-transformed free-text reports. Here, the model should not only learn the granularity of input but also our ReportQL syntax that the output should be structured with.

\section{Results and Experience}
\subsection{Annotation performance}

As mentioned above, 88 abdominopelvic sonography free-text reports were divided into training and test sets (80\%-20\%). The train set was manually annotated by two radiologists using the mentioned information schema and used to train our base NER model. Then, the model was evaluated with the validation set of abdominopelvic sonography reports. An example of the input free-text report with the annotated structured result is shown in table 1. Each input report consists of the description of approximately 17 abdominopelvic organs with an average of 22 fine-grained properties (min: 8, max: 35) in the free-text format which is annotated to extract the mentioned entities and relations and designed in the form of structured data containing the entities related to each organ in the brackets. 

\subsection{Evaluation metrics}

We evaluated the performance of our models using ROUGE (ROUGE-1, ROUGE-2, and ROUGE-L) besides the SacreBLEU and BLEU scores. The Scifive base model has achieved a ROUGE-1 score of 0.732 with subsequent scores of 0.633 and 0.635 for ROUGE-2 and ROUGE-L metrics, respectively. The Scifive-Mask model also achieved ROUGE-1, ROUGE-2, and ROUGE-L scores of 0.723, 0.550, and 0.610.

Regarding BLEU scores \cite{papineni2002bleu}, the Scifive base model has achieved SacreBLEU and BLEU scores of 0.738 and 0.727, respectively. The Scifive-Mask model has shown the same performance with SacreBLEU and BLEU scores of 0.721 and 0.710, respectively with a brevity penalty and length ratio of 0.99. Tables \ref{table:2} and \ref{table:3} have summarized the mentioned metrics.

\setlength{\tabcolsep}{0.8em} 
{\renewcommand{\arraystretch}{1.5}

\section{Discussion}

In this study, we proposed a Scifive-based transfer learning model to annotate free-text radiology reports and provide structured texts interpretable by both physicians and data storage software. The model not only achieved BLEU and ROUGE-L scores of 0.727 and 0.635, but also it can extract entities and relations, identify abbreviations and negation, provide a clinically interpretable structured report, and give the operator freedom on which organs are in the structured report.

Our proposed model has shown significant strengths compared with the previous models. First, it can identify relations of the used entities in the report besides the entity extraction. Although in NLP tasks, NER should be accompanied by relation extraction to convert unstructured free texts to a structured format, most studies have focused only on one of these methods and in particular the NER. In a study conducted by Sugimoto et al., BiLSTM-CRF, BERT, and BERT-CFR models were used for entity extraction. Annotated chest CT scans were used to train the model to extract the terms and categorize them in the observation entity, clinical finding entity, and modifier entity. Not extracting the relations among the identified terms can limit the applicability of the mentioned model that should be solved towards developing structured texts \cite{sugimoto2021extracting}. Therefore, to enable the creation of a structured report in this study, our proposed model has been designed to extract existing relations. Considering the sentence “few stones up to 7 millimeters are seen in the upper pole of the left kidney”, our model provides the result of “stones {quantity {few}, size {up to 7 mm}, and location {upper pole}}” with identified entities and relations.

Second, after extracting entities and their relations, it can use these entities to provide a clinically interpretable structured report that can be used as an alternative to free-text narratives in clinical practice. As mentioned, most of the available NLP experiments have focused on entity extraction that aims to answer a specific question such as the presence of a finding or follow-up indication; however, scant studies have provided an information model capable of transforming reports to structured formats. In a study published by Huang et al. in 2019, an encoder-decoder architecture based on Bidirectional Long Short-Term Memory (Bi-LSTM) and the unidirectional Long Short-Term Memory (LSTM) was applied to annotate chest CT reports. They first annotate the reports as “normal” or “abnormal” and then use NER to annotate 23 possible pathologies found in chest CT scans followed by a complete annotation step. The model can identify normality or abnormality of a report and extract entities presented in the abnormal texts; however, the model cannot annotate normal features and design the abnormal found entities in the form of a structured text to present limiting its clinical applicability \cite{huang2019annotation}.

\begin{table*}[t]
	\begin{center}
		\centering
		\begin{tabular}{l|cccccc}
			\toprule
			\hline
			   & ROUGE-1 & ROUGE-2 & ROUGE-L & BLEU & SacreBLEU\\
			\hline
			Scifive&  0.732 & 0.633 & 0.635 & 0.727 & 0.738\\
			
			Scifive-Mask&  0.723 & 0.550 & 0.610 & 0.710 & 0.721\\
			\hline
			\bottomrule
		\end{tabular}
	\end{center}
	\caption{Summary of the ROUGE-1, ROUGE-2, ROUGE-L, BLEU, and SacreBLEU scores for Scifive and Scifive-Mask models.}
	\label{table:2}
\end{table*}

Besides, heterogeneity of the grammatical formats and terminologies used by different radiologists has limited this process and made the development of encoded structured reports cumbersome. For example, reports may provide some normal or insignificant findings or normal variations of observations instead of a clear normality statement that should not be interpreted as an abnormal feature. Also, negation detection, heterogeneous sentences used to report abnormalities, synonyms, abbreviations, not using common standardized medical terminologies, etc. challenge medical text processing. 

However, our model has shown promising results in these aspects facilitating the development of structured reports. The model has demonstrated significant performance in synonyms detection. As an example, all the phrases “dilated bile ducts”, “biliary tree dilatation”, and “intrahepatic biliary duct dilatation” were correctly detected by the model as synonyms. Commonly used medical abbreviations such as “CBD” (common bile duct), “GB” (gallbladder), “SOL” (space-occupying lesion), etc. were also correctly diagnosed. The model has also shown accurate performance in negation detection and in particular in grammatically complex sentences. Besides, in sentences where the same findings are reported for two organs such as “both kidneys” or “both ovaries”, the model can differentiate and detect both right and left organs separately and mention all the common findings for both of them.

Third, flexibility and lack of word count limitation can also make this model more applicable. The operator can determine which organs will be included in the structured format. For instance, if we aim to collect and analyze characteristics of the liver organ in all the population of an area or patients referring to a specific center, the model can be modified to extract only the information related to the liver organ in all the reports and present them in a structured format to use in the analysis. On the other hand, there is no limitation regarding the number of words and sentences used in the reports compared with previously mentioned models. As mentioned, CheXbert is a BERT-based model for performing automatic labeling of chest X-rays. Despite its accurate performance, the model faces some limitations. First, it can receive a maximum of 512 tokens as the input, and second, it can only identify the presence of 13 predefined radiologic abnormalities and all the other observations would be marked as “no finding”. Besides, the model defines one of the four categories of “positive, negative, uncertain, and blank” for each of these 13 medical terms; however, it cannot provide further information regarding the characteristics of the positive findings \cite{smit2020chexbert}. However, our proposed model has no limitations regarding the size and number of medical terminologies that can label and can also provide further characteristics of each positive observation such as the anatomic and severity modifiers. Besides, the acceptable performance of our model despite using a small corpus of abdominopelvic sonography reports as the train set compared with the other recent studies would highlight the strengths of the model.

In addition, the model performance was also evaluated using BLEU and ROUGE-L scores. Regarding the BLEU score, the Scifive model has achieved a score of 0.727 with almost the same score in the Scifive-Mask model. The ROUGE-1 and ROUGE-L scores were also measured to be 0.732 and 0.635 in the Scifive model. Considering these metrics, our proposed Scifive model has outperformed previous similar models. Mostafiz et al. in 2018 applied a deep neural network (DNN) for NER in radiology reports to extract pathological terms which achieved the BLEU-1 score of 0.49 reflecting better performance in comparison to other medical annotation tools such as IBM-Natural Language Understanding (IBM-NLU) and Medical Text Indexer (MTI) with Non-MEDLINE and Mesh On Demand (MOD) options \cite{mostafiz2018pathology}.

From the modeling perspective, it seems that transformer-based models outperformed RNN and CNN while requiring less time to train the model and can be generalized to a wide range of tasks \cite{vaswani2017attention}.

The model also has some limitations. First, it was only trained based on the abdominopelvic sonography reports performed at two medical clinics in Shiraz, Iran. Testing the model performance with radiology reports obtained from other medical centers and including reports of other imaging modalities such as CT scans and MRI images related to chest, brain, and other organs should be performed in the next steps. Second, although the model has shown perfect performance despite the relatively small corpus of reports used as the train set, more reports should be included to improve the model performance. Third, the model is shown to miss some phrases that are rarely reported in radiology texts such as the description of a rare finding or some suggestion statements. However, this problem would be solved by increasing the input report’s number in further steps.

\begin{table*}[t]
	\begin{center}
		\centering
		\begin{tabular}{l|cccccc}
			\toprule
			\hline
			
			   & Score & Brevity penalty & Length ratio & Precision\\
			\hline
			Scifive &  0.7382 & 1.0 & 1.01 & [0.84, 0.76, 0.69, 0.62]\\
			
			Scifive-Mask &  0.7100 & 0.99 & 0.99 & [0.85, 0.76, 0.68, 0.60]\\
			\hline
			\bottomrule
		\end{tabular}
	\end{center}
	\caption{Brevity penalty, length ratio, and precision metrics used for BLEU score calculation. }
	\label{table:3}
\end{table*}

\section{Conclusion}

Considering the increasing application of transformer-based NLP models in recent years, using a text-to-text transfer transformer model (T5) and a pre-trained domain-specific adaptation of T5 (Scifive) in this study has shown accepted performance for NER of radiology reports and translation of free-text records to the structured data. The model would also handle annotation of other imaging modalities and also pathological and biological reports with minimal changes in the model schema.

\section{Statements and Declarations}

\textbf{Competing interests statement:} The authors declare that they have no conflict of interest.

\begin{table*}[ht]
	\begin{center}
		\fontsize{9}{9.5}\selectfont
		\centering
		\begin{tabular}{p{0.45\linewidth} | p{0.5\textwidth}}
			\hline
			Abdominopelvic sonography free-text report   &	Abdominopelvic sonography structured report equivalent received from our model\\
			\hline
			The liver is normal in size and with normal parenchymal echogenicity with no sign of space-occupying lesion or bile ducts dilatation. GB is well distended with no stone or wall thickening. The spleen is normal in size and parenchymal echogenicity with no sign of space-occupying lesion. visualized parts of the pancreas and para-aortic area are unremarkable. Both kidneys are normal in size with normal cortical parenchymal echogenicity with no sign of the stone, stasis, or perinephric collection. ureters are not dilated. The urinary bladder is empty so evaluation of pelvic organs is not possible. no free fluid is seen in the abdominopelvic cavity. & 
			\mintinline[linenos,tabsize=2,breaklines, gobble=0]{tex}{liver { size { normal } echogenicity { normal } lesion { no } bile duct { no } } GB { seen { yes } distention { well } stone { no } wall thickening { no } } spleen { size { normal } echogenicity { normal } lesion { no } } pancreas { seen { yes } } right kidney { size { normal } cortical parenchymal { normal } stone { no } stasis severity { no } perinephric collection { no } } left kidney { size { normal } cortical parenchymal { normal } stone { no } stasis severity { no } perinephric collection { no } } right ureter { dilation { no } } left ureter { dilation { no } } bladder { distension empty evaluation { yes } } abdominopelvic cavity { free fluid { no } }}
			
			\\
			\hline
			\bottomrule
		\end{tabular}
	\end{center}
	\caption{An example of the free-text abdominopelvic sonography report with the final structured report received from the model.}
	\label{table:1}
\end{table*}

\section{Acknowledgments}

This research did not receive any specific grant from funding agencies in the public, commercial, or not-for-profit sectors. We thank Dr. Sadra Valiee (Shiraz University of Medical Sciences) and Dr. Mohammadjavad Gholamzadeh (Shiraz University of Medical Sciences) for their contribution to providing the information schemas.

\bibliography{bib}
\end{document}